\documentclass[journal,onecolumn]{IEEEtran}
\usepackage{generic}
\usepackage{cite}
\usepackage[T1]{fontenc}
\usepackage{float}
\usepackage{cuted}    
\usepackage{capt-of} 
\usepackage{amsmath,amssymb,amsfonts}
\usepackage{algorithmic}
\usepackage{graphicx}
\usepackage{textcomp}

\begin{document}
\title{Automatic breach detection during spine pedicle drilling based on vibroacoustic sensing}
\author{Aidana Massalimova$\ast$, Maikel Timmermans$\ast$, Nicola Cavalcanti, Daniel Suter, Matthias Seibold, Fabio Carrillo, Christoph J. Laux, Reto Sutter, Mazda Farshad, Kathleen Denis, Philipp Fürnstahl
\thanks{'$\star$' means that the authors contributed equally to this work and share the first authorship. The work was supported by the FAROS project. This project has received funding from the European Union's Horizon 2020 research and innovation program under grant agreement No. 101016985. }
\thanks{ A.Massalimova is with the Research in Orthopedic Computer Science (ROCS), University Hospital Balgrist, University of Zurich, Zurich, Switzerland (e-mail: aidana.massalimova@balgrist.ch).}
\thanks{M.Timmermans is with KU Leuven, Department of Mechanical Engineering, BioMechanics (BMe), Smart Instrumentation Group, Leuven, Belgium (e-mail: maikel.timmermans@kuleuven.be). }
\thanks{N.Cavalcanti is with the Research in Orthopedic Computer Science (ROCS), University Hospital Balgrist, University of Zurich, Zurich, Switzerland (e-mail: nicola.cavalcanti@balgrist.ch). }
\thanks{D.Suter is with the Research in Orthopedic Computer Science (ROCS), University Hospital Balgrist, University of Zurich, Zurich, Switzerland (e-mail: daniel.suter@balgrist.ch). }
\thanks{M. Seibold is with the Research in Orthopedic Computer Science (ROCS), University Hospital Balgrist, University of Zurich, Zurich, Switzerland (e-mail: matthias.seibold@balgrist.ch). }
\thanks{F.Carrillo is with the Research in Orthopedic Computer Science (ROCS), University Hospital Balgrist, University of Zurich, Zurich, Switzerland (e-mail: fabio.carrillo@balgrist.ch).}
\thanks{C.J.Laux is with the Department of Orthopaedics, Balgrist University Hospital, University of Zurich, Zurich, Switzerland (e-mail: christoph.laux@balgrist.ch). }
\thanks{R.Sutter is with the Department of Radiology, Balgrist University Hospital, Zurich, Switzerland (e-mail: reto.sutter@balgrist.ch). }
\thanks{M.Farshad is with the Department of Orthopaedics, Balgrist University Hospital, University of Zurich, Zurich, Switzerland (e-mail: mazda.farshad@balgrist.ch). }
\thanks{K.Denis is with KU Leuven, Department of Mechanical Engineering, BioMechanics (BMe), Smart Instrumentation Group, Leuven, Belgium (e-mail: kathleen.denis@kuleuven.be). }
\thanks{P.Fürnstahl is with the Research in Orthopedic Computer Science (ROCS), University Hospital Balgrist, University of Zurich, Zurich, Switzerland (e-mail: philipp.fuernstahl@balgrist.ch). }
}

\maketitle

\begin{abstract}
Pedicle drilling is a complex and critical spinal surgery task. Detecting breach or penetration of the surgical tool to the cortical wall during pilot-hole drilling is essential to avoid damage to vital anatomical structures adjacent to the pedicle, such as the spinal cord, blood vessels, and nerves. Currently, the guidance of pedicle drilling is done using image-guided methods that are radiation intensive and limited to the preoperative information. This work proposes a new radiation-free breach detection algorithm leveraging a non-visual sensor setup in combination with deep learning approach. Multiple vibroacoustic sensors, such as a contact microphone, a free-field microphone, a tri-axial accelerometer, a uni-axial accelerometer, and an optical tracking system were integrated into the setup. Data were collected on four cadaveric human spines, ranging from L5 to T10. An experienced spine surgeon drilled the pedicles relying on optical navigation. A new automatic labeling method based on the tracking data was introduced. Labeled data was subsequently fed to the network in mel-spectrograms, classifying the data into breach and non-breach. Different sensor types, sensor positioning, and their combinations were evaluated. The best results in breach recall for individual sensors could be achieved using contact microphones attached to the dorsal skin (85.8\%) and uni-axial accelerometers clamped to the spinous process of the drilled vertebra (81.0\%). The best-performing data fusion model combined the latter two sensors with a breach recall of 98\%. The proposed method shows the great potential of non-visual sensor fusion for avoiding screw misplacement and accidental bone breaches during pedicle drilling and could be extended to further surgical applications.
\end{abstract}


\section{Introduction}
\label{sec:introduction}
Pedicle screw placement is a standard procedure in spinal surgeries performed in cases of scoliosis, trauma, spinal tumors, and degenerative spinal pathologies. However, due to adjacent vital anatomical structures, sub-optimal pedicle screw positioning can lead to neurological and vascular injuries. It was reported that the rate of pedicle screw malpositioning reaches up to 41\% in the lumbar spine and 55\% in the thoracic spine \cite{perna2016pedicle}.
The optimal screw positioning is achieved when screws have maximum diameter and length within the pedicle, do not breach the pedicle's cortical layer or vertebral body, and follow a converging trajectory usually parallel to the upper endplate. In conventional freehand techniques, a profound understanding of the anatomical landmarks is essential to identify the entry point. After finding the entry point, a surgical drill or a probe is used to create the pilot hole inside the pedicle. This technique demands thorough anatomical knowledge and experience-based judgment from tactile feedback to avoid pedicle wall perforation.
\par
With the advances in imaging and robotic technologies, imaging-based and robot-assisted techniques were developed to navigate the surgeon during pedicle screw positioning and prevent pedicle breaches. In \cite{gelalis2012accuracy}, authors performed a meta-analysis and reported 28–85\%, 81–92\%, and 89–100\% of screws fully contained within the pedicle for freehand, fluoroscopy-guided and marker-based optical navigation, respectively. However, the fluoroscopy-guided method is characterized by higher radiation exposure \cite{perna2016pedicle,dusad2018}. A systematic study juxtaposing the robot-assisted and freehand methods proved improved accuracy, reduced radiation exposure, and prolonged operation time with TiRobot (TINAVI Medical Technologies, Beijing, China)\cite{peng2020accuracy}. Nevertheless, this system is based on marker-based optical navigation and preoperative planning, similar to imaging-guided techniques. Preoperative planning is performed in dedicated software that allows defining the drilling trajectory using 3D anatomy models segmented from preoperative imaging, e.g., computed tomography (CT). However, a non-negligible patient movement during the operation introduces additional errors to the system. These issues were tackled by a smart commercial surgical instrument called PediGuard Probe (SpineGuard, Vincennes, France). It measures the local electrical conductivity of the tissue in contact with the probe tip and, based on that, emits a sound that varies in pitch and cadence according to the changes in the nature of the bone or tissue. \cite{chaput2012reduction} compared the PediGuard Probe with the fluoroscopy-assisted method and demonstrated that both methods have an accuracy of 97.5\%. However, the number of fluoroscopy shots was reduced by 30\% with the aid of the PediGuard probe. \cite{Bolger2007} found a similar accuracy of >98\% with the use of fluoroscopic guidance, whereas \cite{guillen2014independent} tested the PediGuard probe without any fluoroscopic guidance and achieved 90.06\% accuracy.  \par
Previously, we conducted a systematic study on the existing intraoperative tissue classification methods in orthopedic and neurological studies  \cite{massalimova2022intraoperative}, where we reported the potential of sensor fusion and vibroacoustic sensing with transducers such as microphones and accelerometers in differentiating cortical and cancellous bone layers. Since pedicle breach occurs in the intersection between cortical and cancellous layers, vibroacoustic sensing can predict pedicle breaches during drilling \cite{massalimova2022intraoperative}. Besides, the non-visual nature of the vibroacoustic sensing alleviates the limitation to the preoperative planning and exhibits less radiation exposure than the methods above. The use of microphones has been extensively investigated for this purpose~\cite{zakeri2017classifying,Guan2018,Sun2014,Torun2018,torun2020parametric}. \cite{Sun2014} designed an algorithm based on the energy in the acoustic emission signal from a 'sound sensor' to distinguish between drilling in cortical or cancellous bone by defining an energy threshold. The signal was ideally analyzed in the range of 8 to 14 kHz, and drilling state recognition rates between 65 and 89\% were found with the best results in fresh bones. \cite{Guan2018} performed similar research with the same sensor but trained a backpropagation neural network for bone layer identification that reached an accuracy of 84\%. In \cite{torun2020parametric, Torun2018}, authors analyzed the sequential frequency change in spectrograms to detect the starting point of every new drilling state. The breakthrough detection algorithm monitored the mean and median frequency in the signal. \cite{zakeri2017classifying} used a microphone to capture free-field sound while drilling a bovine tibia. The spectrogram of each drilling process was divided into five frequency bands, separately used as input to a support vector machine algorithm. A subject-specific approach was used as the classifier was consistently primed with the data of the first cortical layer at the start of the drilling process to enhance the classification performance for the second cortical layer in the continuation of the drilling process. This approach proved more accurate than generalizing classification across bone specimens, as the average total accuracy rose from 70.9 to 83\% when the data of the first cortical layer of the respective sample was used to prime the classifier. \par
A mel-spectrogram is a visual representation of the audio signal's amplitude over time across different frequency bins in the mel scale. The latter is a scale proportional to the perceived (by humans) difference in magnitude of different frequencies at the same magnitude due to the cochlear anatomy~\cite{stevens1937scale}. Recently, \cite{seibold2021real} showed that a convolutional neural network, using mel-spectrograms as input, can detect the breakthrough event during a femur drilling process with an accuracy of over 90\% in a total execution time of less than 140 ms. Inspired by this method, we propose a novel method of detecting pedicle breaches during the pedicle drilling step in spine surgery by leveraging the power of vibroacoustic sensing and convolutional neural networks (CNNs). A breach event is considered as the penetration of the surgical instrument from the cancellous layer to the adjacent cortical of the vertebra. This study aims to analyze the performance of the CNN with different vibroacoustic sensors and to investigate whether and to which extent data fusion of those different sensors improves the breach detection rate during pedicle drilling. We hypothesize that the performance of the CNN can be boosted in the case of sensor fusion since they might provide complementary information. The following contributions can substantiate the novelty of this study:
\begin{enumerate}
    \item Our work introduces an automatic labeling approach based on tracking data from the optical navigation system for non-visual signal data that greatly accelerates data collection for machine learning.
    \item Multiple vibroacoustic sensors, including a contact microphone, a free-field microphone, a tri-axial accelerometer, and a uni-axial
    accelerometer, were integrated into the setup such that we could systematically investigate vibroacoustic sensing technologies both individually and fused. 
    \item A breach detection algorithm was developed by leveraging sensor fusion and a squeeze-and-excitation neural network.
    \item The data collection was performed by conducting a series of ex-vivo cadaveric experiments to facilitate close to the actual clinical application scenario.
    \item The concept of transfer learning in the mel-spectrogram domain
    was tested for the first time using pre-trained weights from a previous study and re-training the network on vibroacoustic sensor data.

\end{enumerate}

\section{Methods}
\label{sec:methods}
\begin{figure*}[ht]
    \centering
    \includegraphics[width=\linewidth]{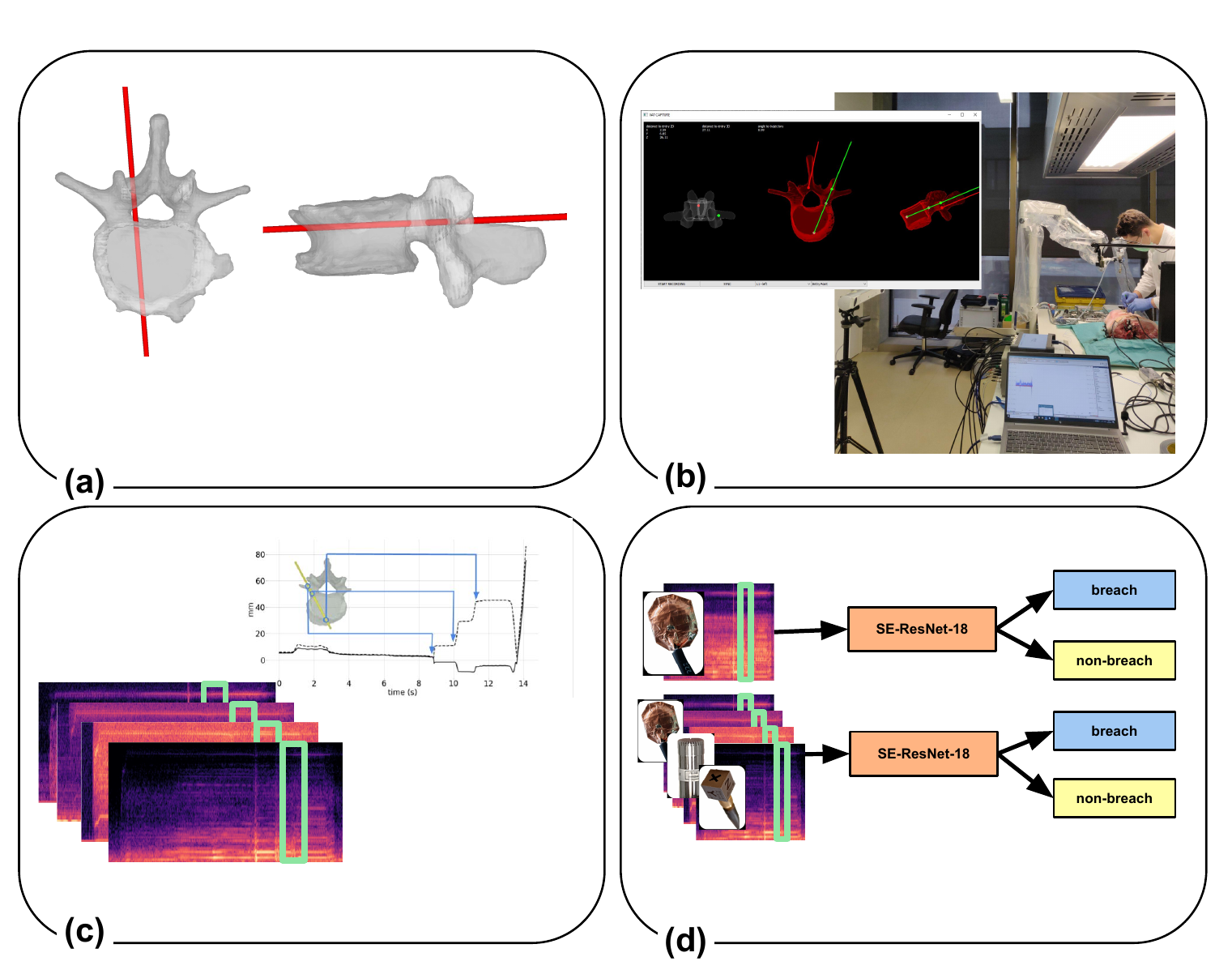}
    \caption{Overall pipeline of the proposed method for breach detection during pedicle drilling: a) definition of drilling trajectories in dedicated software, b) drilling by experienced surgeon and data collection, c) processing of the vibroacoustic data and the optical tracking data, d) training the deep learning model on individual sensor data and fused sensor combination data.}
    \label{fig:pipeline}
\end{figure*}

The proposed method includes a definition of drilling trajectories, data collection, data processing, and deep learning, as shown in Fig \ref{fig:pipeline}. Definition of drilling trajectories was performed first as described in section \ref{sec: preoperative planning} to simulate different breaches during pedicle drilling,  which was later visualized in the surgical navigation setup (section \ref{subsubsec: navigation setup}). Vibroacoustic data from contact microphones (\textbf{$Mic_1$}), a free-field microphone (\textbf{$Mic_2$}), uni-axial accelerometers (\textbf{$PCB_1$} and \textbf{$PCB_2$}) and a tri-axial accelerometer (\textbf{$PCB_3$}) were collected during pedicle drilling along with surgical navigation data (section \ref{subsec: data collection}). Collected data were processed (section \ref{subsec: data processing}) and used to develop a CNN for breach detection in section \ref{subsec:deep learning}.
\subsection{Definition of Drilling Trajectories}
\label{sec: preoperative planning}
\begin{figure}[ht]
    \centering
    \includegraphics[scale =0.6]{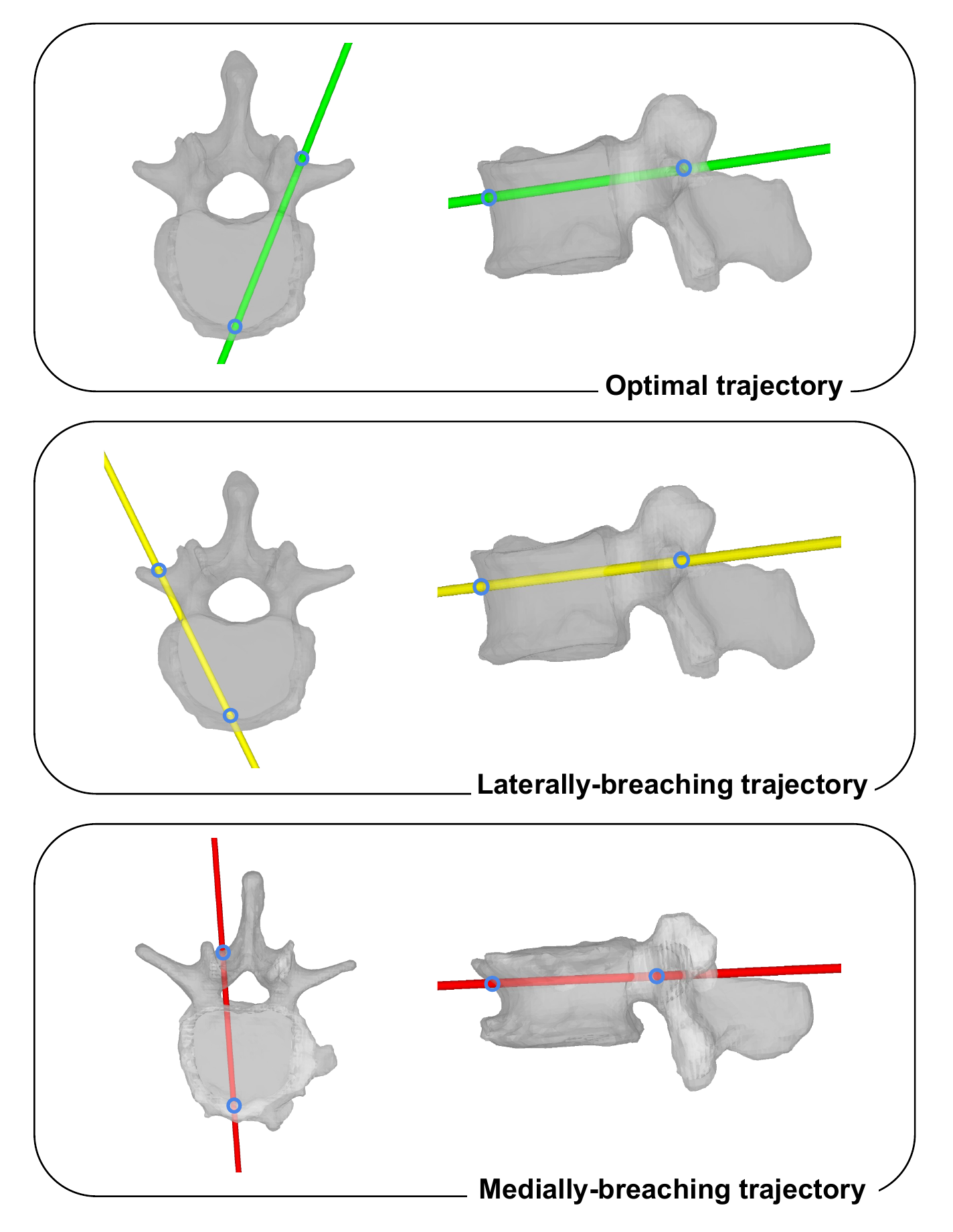}\\
    \caption{ Optimal, laterally-breaching, and medially-breaching trajectories overlaid on 3D meshes of the vertebra in axial and sagittal views.}
    \label{fig:planning}
\end{figure}

Four fresh-frozen human cadaver spines from T12 to coccyx were acquired from Science Care (Phoenix, USA) and used for data collection. Ethical approval was obtained from the ethical committee for conducting this cadaveric study (ID: 2021-01196). The spines were fixed onto a wooden board using surgical pins with a diameter of 3 mm to provide stability to the anatomy during experiments. The soft tissue on top of the sacrum was removed. A marker with four converging canals was designed, 3D-printed, and attached to the sacrum of each specimen to track relative motion between anatomy and marker. CT scans of each specimen with a marker and infrared reflecting spheres were taken with a slice thickness of 1 mm (SOMATOM Edge Plus, Siemens Healthcare, Erlangen, Germany). Afterward, all vertebrae, markers, and spheres were segmented and converted into 3D triangular surface models using the Materialise Mimics Innovation Suite software  (version 19.0, Materialise NV, Leuven, Belgium). Drilling trajectories were planned on each spine, from T10 to L5 levels, and on both sides of each vertebra using the surgery planning tool of our institution. Trajectories were planned by resident surgeons (NC and DS) and confirmed by an experienced spine surgeon (CL). Along with optimal or breach-free trajectories, laterally- and medially-breaching trajectories were planned as shown in Fig. \ref{fig:planning}. In all cases, the anterior cortical wall of the vertebral body was deliberately breached.

Cadavers were thawed a day previous to the experiment, and the surgical approach was performed by a surgeon, providing midline access to the posterior elements of the thoracolumbar spine. 
\subsection{Data Collection}
\label{subsec: data collection}
\subsubsection{Sensors}
\label{subsubsec:sensors}
\begin{figure*}
    \centering
    \includegraphics[scale = 0.42]{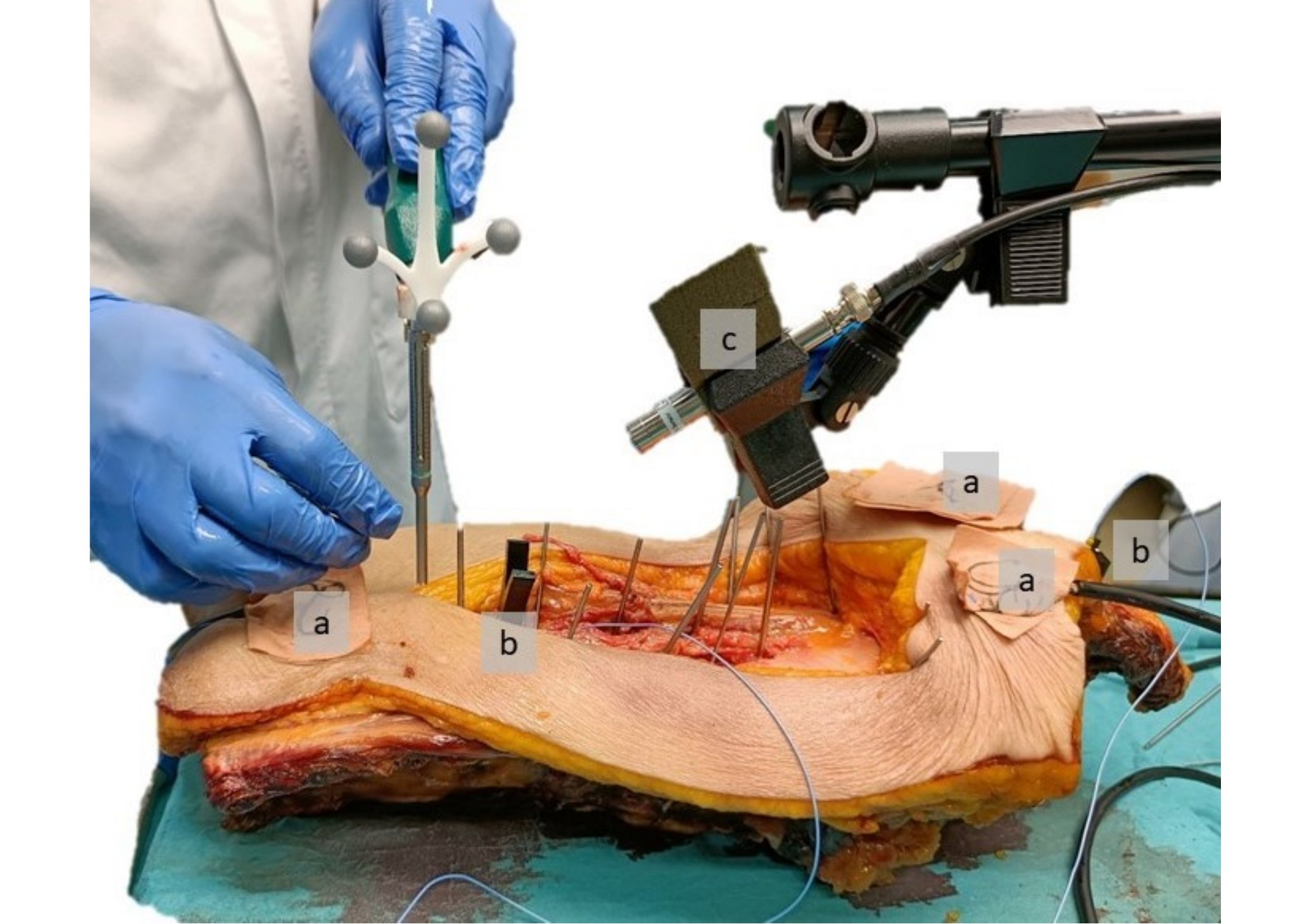}
    \includegraphics[scale = 0.40]{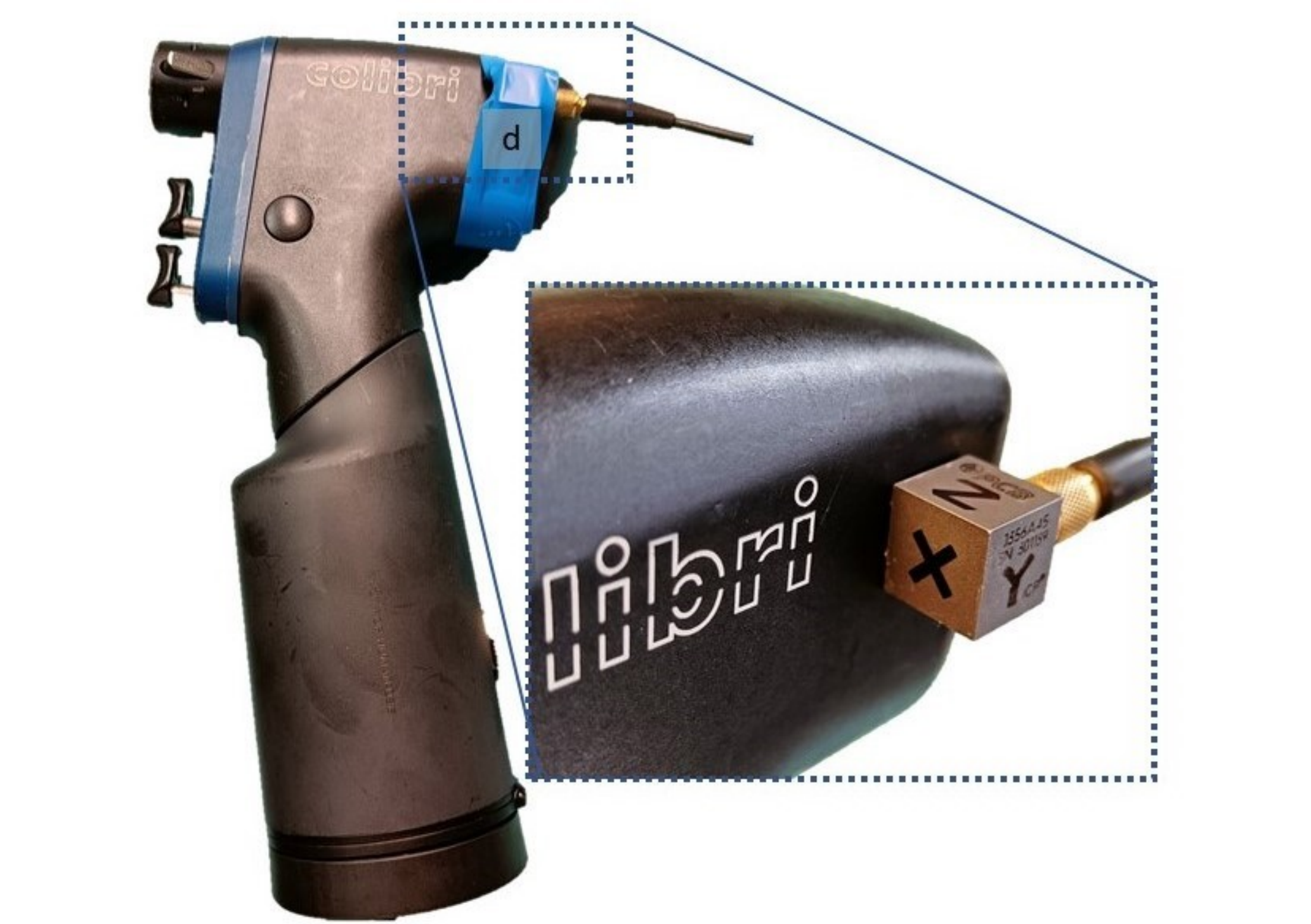}
    \caption{Sensors positioning. Left: (a) contact microphones, (b) uni-axial accelerometers, and (c)  free-field microphones. Right: (d) tri-axial accelerometer.}
    \label{fig:setup}
\end{figure*}
Fig.~\ref{fig:setup} shows all sensors in contact with the sample, the free-field microphone above the sample on the left, and the tri-axial accelerometer attached to the drill on the right. Three custom piezo-electric contact microphones ($Mic_1$) were attached to the skin directly above bony structures near the incision site using kinesiology tape and connected to an analog/digital converter (PreSonus Studio 68, PreSonus Electronics, Inc., Baton Rouge, LA, USA) through a 48V phantom-powered impedance matching circuit\footnote{designed by Alex Rice: https://www.zachpoff.com/resources/alex-rice-piezo-preamplifier/}, to tackle the impedance mismatch issue (Fig.~\ref{fig:setup}a). A PCB free-field condenser microphone ($Mic_2$) of model 378B02 (PCB Piezotronics, Depew, NY,  USA) was placed at a distance of approximately ten to twenty centimeters from the drilling site using a standard tripod to capture acoustic variations in bone properties similar to~\cite{Goossens2020} as shown in Fig.~\ref{fig:setup}c. One PCB uni-axial lightweight accelerometer ($PCB_1$) of model 352A24 (PCB Piezotronics, Depew, NY,  USA) was attached to the spinous process of the vertebra being drilled, utilizing an in-house manufactured 3D-printed clamp (Fig.~\ref{fig:setup}b). Another uni-axial accelerometer ($PCB_2$) of the same model was attached to a surgical pin (Fig.~\ref{fig:setup}b), and drilled through the sacral bone to fix the spine to the wooden board. On the surgical power drill, a tri-axial accelerometer ($PCB_3$) of model J356A45 (PCB Piezotronics, Depew, NY, USA) was glued with beeswax and taped over to prevent accidental detachment (Fig.\ref{fig:setup}d). All PCB sensors were connected to a custom signal conditioner via BNC cables, and the signal was put through to a PicoScope 4824A oscilloscope (Pico Technology, Cambridgeshire, UK) for data recording. 
\par
The recording of all sensors was started, and the surgeon performed the entire drilling procedure for the selected pedicle at the maximum drilling speed of 3500 rpm. The surgeon started the drilling process guided by the surgical navigation system, as described in the next section. Post-operative CT scans were analogous to the preoperative CT scans after each experiment to compare the drilled trajectories marked with k-wires with the planned trajectories. 
 \\
\subsubsection{Navigation Setup}
\label{subsubsec: navigation setup}
\begin{figure*}
    \centering
    \includegraphics[scale = 0.6]{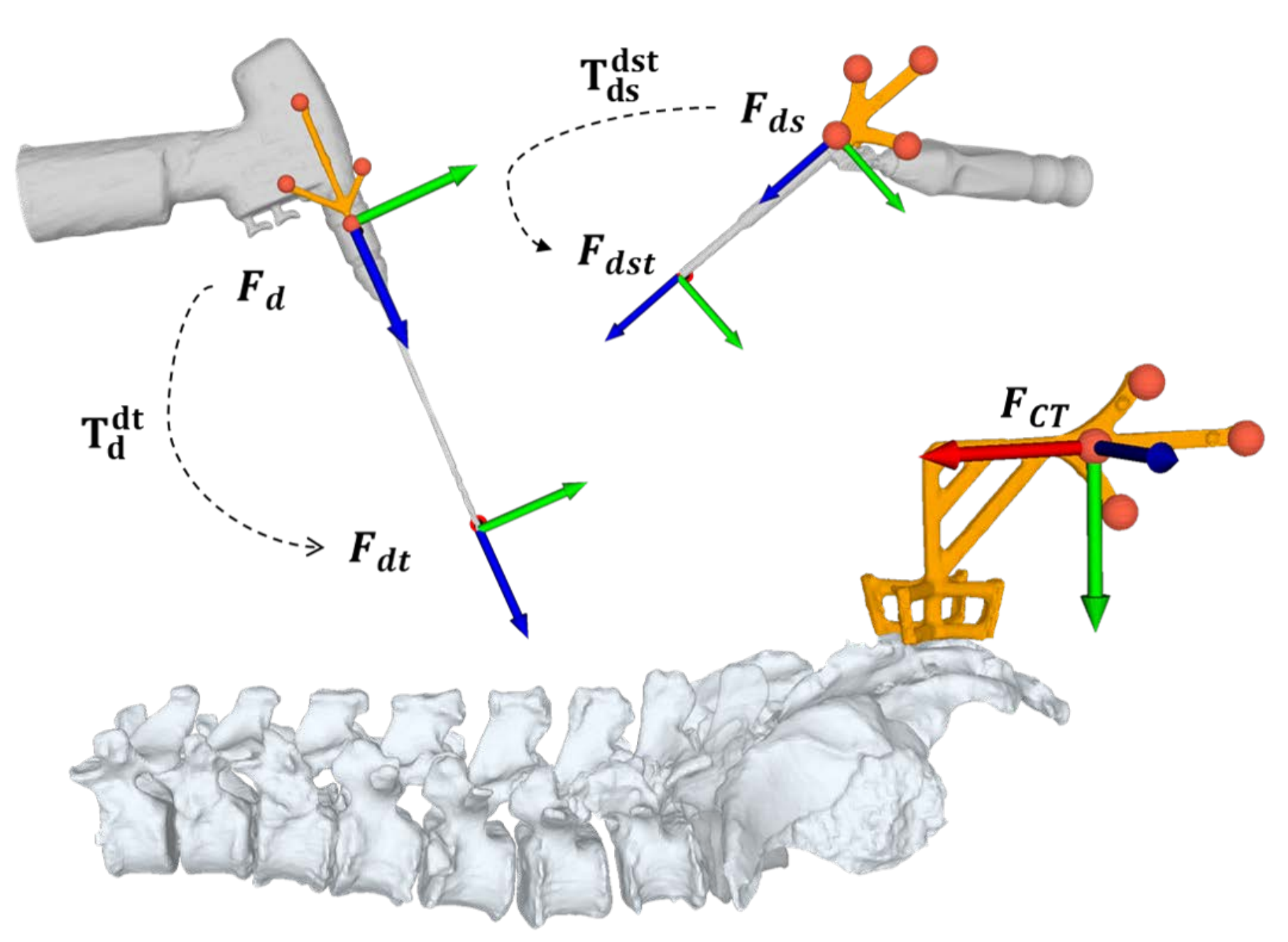}
    \caption{Navigation setup overview with frames (F) and transformations (T). $F_d$ and $F_{ds}$ are the frames of markers attached to the drill and drill sleeve. $F_{ct}$ is a marker frame attached to the anatomy representing CT space. $F_{dt}$ and $F_{dst}$ are tip frames of each tool, which are defined by applying respective transformation matrices (\textbf{$T_d^{dt}$}, \textbf{$T_{ds}^{dst}$}).}
    \label{fig:frames}
\end{figure*}

An optical tracking system (FusionTrack 500, Atracsys LLC, Puidoux, Switzerland) was used to precisely navigate the planned screw trajectories. Similar to the markers designed for the anatomy, markers for the surgical drill sleeve and for a power drill (Colibri, Depuy Synthes, Oberdorf, Switzerland) were modeled. The diameter of the drill bit was 3mm. The drill sleeve was used to find the entry point and direction of the planned trajectory because it provides easier targeting and stable drilling. At the same time, the drill marker allowed for tracking the progression of the drill in depth. Fig. \ref{fig:frames} depicts the designed markers for the drill, drill sleeve, sacrum, and their frames ($F_d$, $F_{ds}$ and $F_{ct}$), respectively. Markers were CT scanned with the infrared reflecting spheres attached. Each tool, sphere, and marker were segmented from preoperative CT scans similarly to the anatomy segmentation, and a 3D mesh per each object was generated. The segmented tools with spheres were used to extract the transformation matrices ($T_d^{dt}$,  $T_{ds}^{dst}$) shown in Fig. \ref{fig:frames} from the marker space to the corresponding tooltip space. Each marker was re-calibrated with FusionTrack 500 software to generate the new geometry file for the tracking. Geometry files contain information on the spatial position of each fiducial in the camera coordinate space, which was used to register 3D meshes of the anatomy or tool to the camera coordinate space using the iterative closest point (ICP) registration method for surgical navigation in the custom-developed software. The software is python-based and allows visualization of the tool position and orientation in real-time with planned entry points and trajectories on 3D meshes of the vertebra. A GUI has been implemented to show the three anatomical views (coronal, axial, and sagittal) of the drilled vertebra (Fig. \ref{fig:pipeline}b). In this manner, the surgeon could meticulously control the drilling procedure such that the drill is positioned in the correct location (i.e., the entry point at the pedicle) and orientation that would result in the planned drilling path. Moreover, the software can record tracking and vibroacoustic data in a synchronized fashion. 
\subsection{Data Processing}
\label{subsec: data processing}
\begin{figure}[ht]
    \centering
    \includegraphics[width = \linewidth]{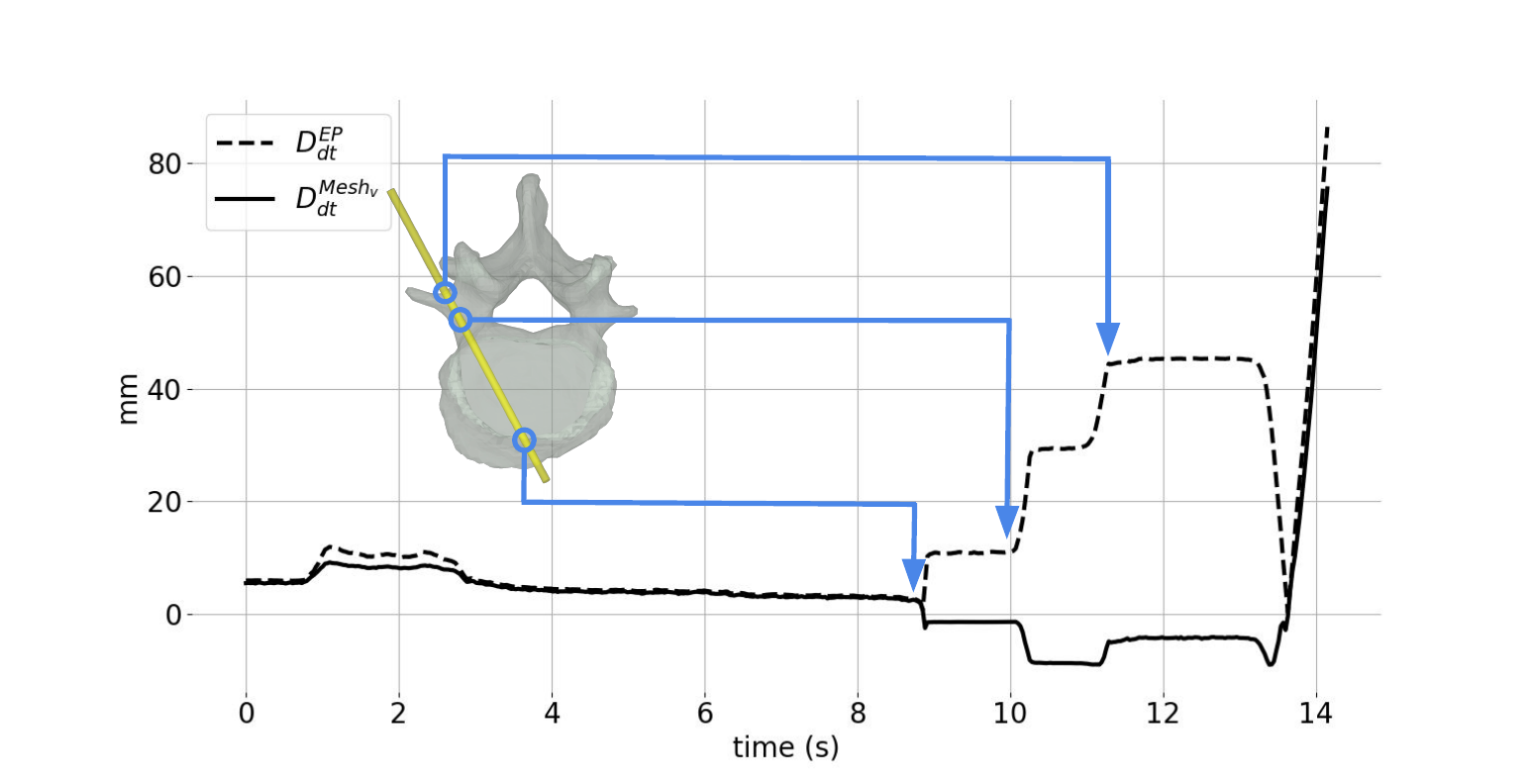}
    \caption{ The blue lines show the correspondence of the start, breach, and stop events between a drill position in a segmented vertebra from postoperative CT and temporal tracking data information. $D_{dt}^{EP}$ is the distance between the drill tip position and the entry point, whereas $D_{dt}^{Mesh_v}$ is the distance between the drill tip position and the closest vertex of the vertebra mesh.}
    \label{fig:tracking_data}
\end{figure}

\begin{figure*}
\centering
\includegraphics[width = 0.9\linewidth]{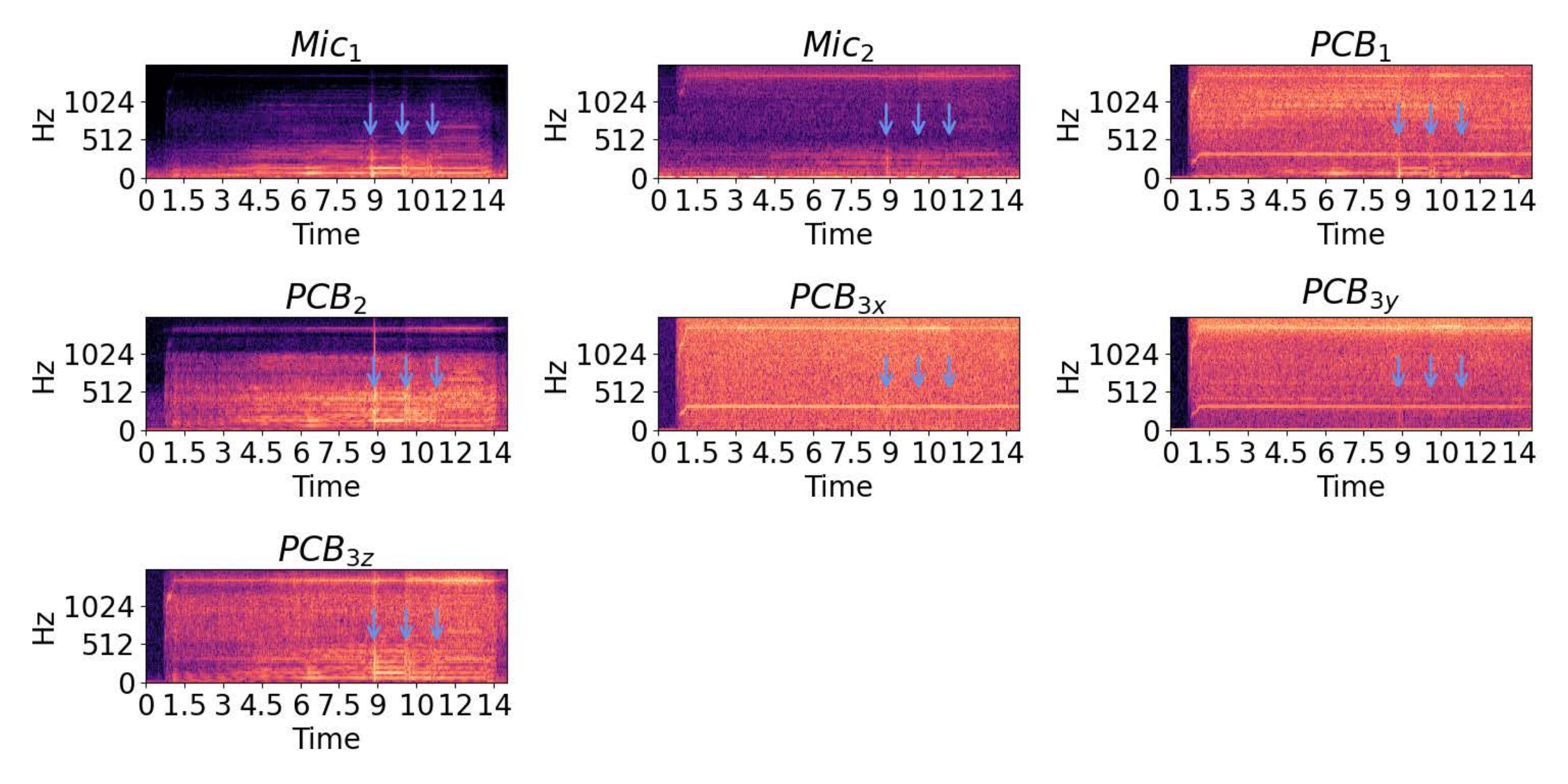}
\captionof{figure}{Mel-spectograms for the whole drilling session from each sensor overlaid with labels from tracking data (Subject S2, vertebra L4, left pedicle). ${PCB}_{3x}$, ${PCB}_{3y}$, and ${PCB}_{3z}$ are separated x (axial),y (radial), and z-directional (tangential) values from the tri-axial accelerometer on the drill. ($PCB_3$).}
\label{Fig: spectograms}
\end{figure*}

Since all sensor modalities are synchronized, data labeling was performed based on the timestamps of the events retrieved from the tracking data. Timestamps were determined for each event to create two data subsets for each sensor: \textbf{breach} and \textbf{non-breach}. A breach event has a duration within a range of 100-300 ms. Following this concept, the closest drill tip positions to the vertebra mesh vertices were used to identify breach timestamps from the tracking data. First, the tracking data was mapped as follows to compensate for non-negligible errors resulting from the navigation setup. A postoperative CT was registered to the preoperative CT scan using ICP registration. Recorded tracking data from the anatomy marker was used to transform 3D meshes from postoperative CT and estimate the distances between the drill tip position and 3D mesh of the vertebra. Entry ($EP_{CT}$) and exit ($SP_{CT}$) points in the CT space were extracted by finding the intersection points between the vertebra and the surgical pin. Afterward, the entry point ($EP$) and exit point ($SP$) in the tracking data of the drill tip were estimated by following Equations \ref{eq1}-\ref{eq2}, where $P_{dt}$ is a set of drill tip position in camera space, $T_{CT}^{cam}$ is a transformation matrix from CT space in $F_CT$ to camera space.
 \begin{equation}
 \label{eq1}
     EP =\underset{P_{dt}}{\arg\min}(||P_{dt}-EP_{CT}T_{CT}^{cam}||)
 \end{equation}
 \begin{equation}
  \label{eq2}
     SP = \underset{P_{dt}}{\arg\min}(||P_{dt}-SP_{CT}T_{CT}^{cam}||)
     \end{equation}
These two points were further used to map the tracked drill tip positions between entering and exiting timestamps to the straight line connecting the entry and exit points (Equation \ref{eq3}). 
 \begin{equation}
    \label{eq3}
     P_{dt} \mapsto \langle P_{dt}[t_{EP}:t_{SP}],EP-SP \rangle
 \end{equation}
To find the breaching point $P_{breach}$, the mapped data was used to find the closest points to the mesh vertices of the vertebra ($Mesh_v$) by transforming it to the camera space using $T_{CT}^{cam}$ as in Equation \ref{eq4}.
 \begin{equation}
   \label{eq4}
    P_{breach} = \underset{P_{dt}}{\arg\min}(||P_{dt}-Mesh_{v}T_{CT}^{cam}||)
 \end{equation}
 The entry, exit, and breaching points are visually illustrated in Fig. \ref{fig:tracking_data} along with their corresponding points on the plots of distance from the drill tip position to the entry point and to mesh vertices in camera space. This visualization of the tracking data was also used to fine-tune timestamp labeling.
On top of that, recorded tracking data was replayed, and timestamps of the breaching events were visually inspected thoroughly. For example, in Fig. \ref{Fig: spectograms}, labels from Fig. \ref{fig:tracking_data} were overlaid in the same time points on mel-spectrograms from each vibroacoustic sensor using blue arrows.

Data from all sensors were processed similarly to keep the data format consistent across vibroacoustic sensors. Each recording per each sensor was broken down into smaller windows using the sliding window. The window length of 100 ms and window step size of 25 ms, leading to overlapping windows with an overlap of 75\%, was chosen to facilitate lower latency. The number of mels, hop size, and maximum frequency included in the spectrogram were set to 128, 32 samples, and 2 kHz, respectively, to generate mel-spectrograms. Table \ref{Table:data_split} shows the number of data per class. 

\begin{table*}
\centering
\caption{Data split per subject and per class before augmentation in each sensor dataset.}
\begin{tabular}{ccccccc}
\hline
\multicolumn{1}{c}{}& \multicolumn{2}{c}{train} & \multicolumn{2}{c}{val}  & \multicolumn{2}{c}{test}                        \\
\hline
\multicolumn{1}{c}{Subject}                                  & \multicolumn{1}{l}{breach} & \multicolumn{1}{l}{non-breach}  & \multicolumn{1}{l}{breach} & \multicolumn{1}{l}{non-breach}  & \multicolumn{1}{l}{breach} & non-breach \\ \hline

\multicolumn{1}{c}{S1}                       & \multicolumn{1}{c}{16}              & \multicolumn{1}{c}{314}                  & \multicolumn{1}{c}{5}               & \multicolumn{1}{c}{78}                  & \multicolumn{1}{c}{3}               & 45                  \\ \hline
\multicolumn{1}{c}{S2}                       & \multicolumn{1}{c}{54}              & \multicolumn{1}{c}{469}                 & \multicolumn{1}{c}{14}              & \multicolumn{1}{c}{117}                 & \multicolumn{1}{c}{9}               & 66                  \\ \hline
\multicolumn{1}{c}{S3}                     & \multicolumn{1}{c}{32}              & \multicolumn{1}{c}{392}                 & \multicolumn{1}{c}{8}               & \multicolumn{1}{c}{98}                  & \multicolumn{1}{c}{5}               & 55                  \\ \hline
\multicolumn{1}{c}{S4}                       & \multicolumn{1}{c}{20}              & \multicolumn{1}{c}{234}                 & \multicolumn{1}{c}{5}               & \multicolumn{1}{c}{58}                  & \multicolumn{1}{c}{4}               & 34                  \\ \hline
\end{tabular}

\label{Table:data_split}
\end{table*}

However, the number of breach samples was considerably smaller, necessitating the application of data augmentation strategies. Pitch shifting and loudness changing are the standard audio signal transformations used for audio augmentation purposes \cite{seibold2021real,schluter2015exploring}. Therefore, the training set of the breach class was augmented by varying the gain (-5 dB, -3 dB, 3 dB, 5 dB) and pitch shifting (-2, -1, 1, 2 semitones). All spectrograms were generated using \textit{librosa 0.9.2} and have a size of 128x126x1. They were further used as input to the deep learning network.

From visual inspection of CT scans, we noticed differences in 
bone mineral density (BMD). Therefore, we estimated BMD from the preoperative CT scans to investigate the correlation between BMD and the performance of the proposed method using the Materialise Mimics software. No density calibration was done for the CT images, but all preoperative scans were taken consecutively to minimize deviations. A constant elliptical area of 450mm² was selected, from which the average Hounsfield unit (HU) was calculated. For each drilled vertebra, the area was swept through all axial CT slices of the vertebral body. All calculated HU values were averaged over the number of axial slices within the vertebral body to estimate the average BMD of the specified vertebra. In~\cite{schreiber2011hounsfield}, HU values calculated from CT scans with the Mimics software were correlated to T-scores generated from DEXA scans. Based on the results of \cite{schreiber2011hounsfield}, the HU values obtained in the present study were related to a value of the BMD, a range of T-scores, and a qualitative description by interpolation on the trend lines. Afterward, test sets were split into either normal or abnormal in the case of osteopenia or osteoporosis, and the data was classified as such (see Table \ref{Table:test bmd}). 

\begin{table}
\centering
\caption{Test split describing the number of the normal and abnormal samples.}
\begin{tabular}{ccccc}
\hline
\multicolumn{1}{c}{} & \multicolumn{2}{c}{normal}                                             & \multicolumn{2}{c}{abnormal}                                                                    \\
\hline
\multicolumn{1}{c}{Subject}                                  & \multicolumn{1}{l}{breach} & \multicolumn{1}{l}{non-breach}  & \multicolumn{1}{l}{breach} & \multicolumn{1}{l}{non-breach}   \\ \hline

\multicolumn{1}{c}{S1}                       & \multicolumn{1}{c}{2}              & \multicolumn{1}{c}{7}                  & \multicolumn{1}{c}{1}               & \multicolumn{1}{c}{38}                     \\ \hline
\multicolumn{1}{c}{S2}                       & \multicolumn{1}{c}{5}              & \multicolumn{1}{c}{18}                 & \multicolumn{1}{c}{4}              & \multicolumn{1}{c}{48}                   \\ \hline
\multicolumn{1}{c}{S3}                     & \multicolumn{1}{c}{0}              & \multicolumn{1}{c}{5}                 & \multicolumn{1}{c}{5}               & \multicolumn{1}{c}{50}                        \\ \hline
\multicolumn{1}{c}{S4}                       & \multicolumn{1}{c}{1}              & \multicolumn{1}{c}{9}                 & \multicolumn{1}{c}{3}               & \multicolumn{1}{c}{25}                   \\ \hline
\end{tabular}
\label{Table:test bmd}
\end{table}

\subsection{Deep Learning}
\label{subsec:deep learning}
\begin{figure*}[ht]
    \centering
    \includegraphics[width = 0.8\linewidth]{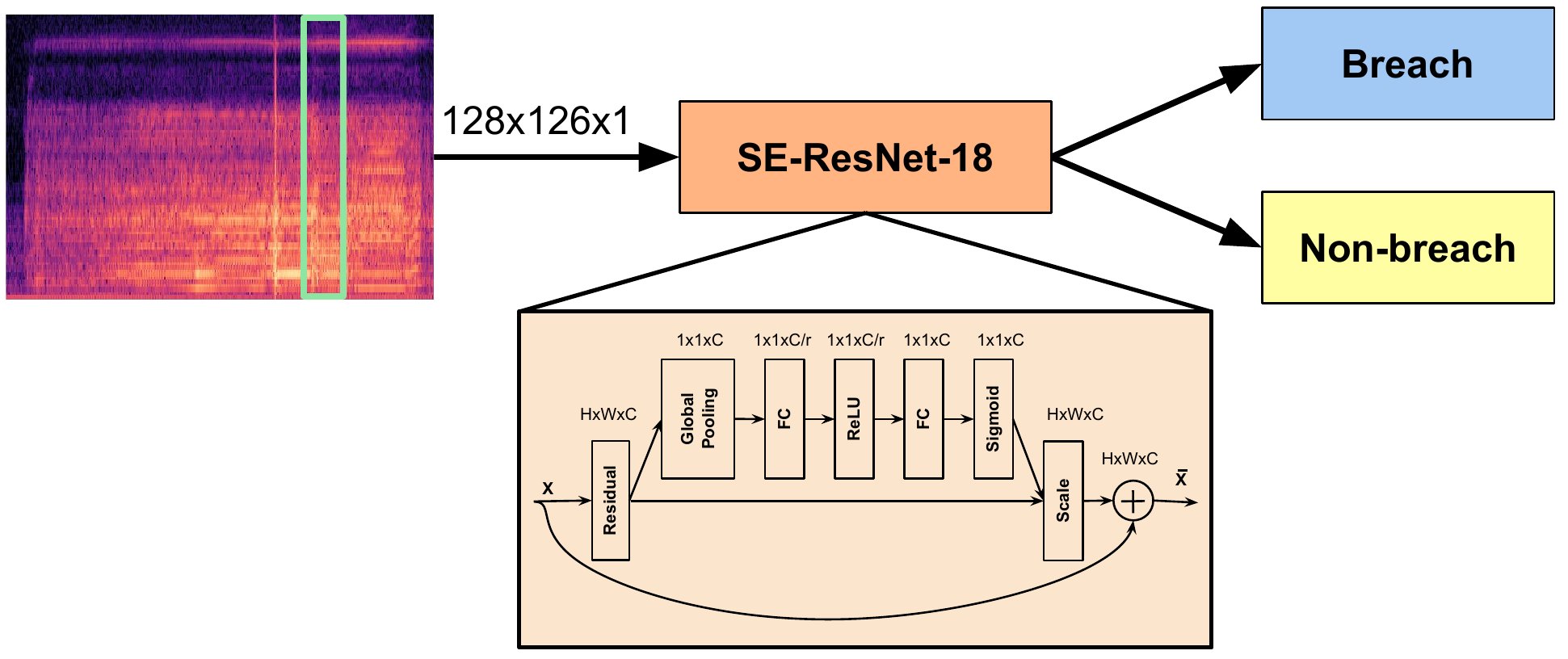}
    \caption{Mel-spectrogram with a window size (green box) of 100ms is an input to the SE-ResNet-18, which outputs breach/non-breach. The topology of the squeeze and excitation block is shown in the brown box. }
    \label{fig:model}
\end{figure*}

This work used a residual network integrated with squeeze and excitation (SE) blocks, particularly SE-ResNet-18 as shown in Fig. \ref{fig:model}. SE blocks enhance the performance of the ResNet networks widely used for classification tasks owing to their ability to model the inter-dependencies between the channels of convolutional features. The SE blocks can be integrated into standard architectures such as ResNet. \cite{hu2018squeeze} reported that ResNet-50 with SE blocks (SE-ResNet-50) outperformed ResNet-50 in different classification tasks, e.g., by 0.86\% single-crop top-5 error with ImageNet 2012 dataset. Since we aim to develop a robust and real-time system, the most compact model of ResNet with 18 layers was used to achieve the fastest inference time. Because of the dataset imbalance, we used the Focal Loss function as a loss function \cite{seibold2021real}.
Further, the Adam optimizer was used with a learning rate of 1e-6. The training and testing were performed on an NVIDIA GeForce RTX 6000 GPU machine using \textit{Tensorflow 2.4.1}. In this work, we used a nested cross-validation method to overcome the problem of overfitting the training dataset. We kept 10\% of the dataset for testing, and the remaining 90\% was divided into 5-folds. As mentioned in the previous section, each experiment was run five times per cross-validation. Thus, results are reported in terms of mean, standard deviation, and 95\% confidence interval (CI) of breach recall (sensitivity).

We ran three different experiments: training the model from scratch based on individual sensor data (Experiment I), training the model using pre-trained weights (Experiment II), and training the model using fused data from different sensor combinations (Experiment III). In Experiment I, we additionally tested with the data from Table \ref{table:bmd test} to investigate whether the performance of each network is correlated with BMD. The flow diagram of the model is shown in Fig.~\ref{fig:model}. Experiment II was based on the pre-trained weights using the same model generated based on the data from \cite{seibold2021real}, and targeted to test the concept of transfer learning from the same domain. Experiments I and II were ablation studies that served as a foundation for finding the optimal sensor combinations in Experiment III, where the spectrograms from different varieties of sensors were concatenated or fused and used as an input to the network (Fig. \ref{fig:pipeline}d). The sensor combinations were formed by combining the two best-performing sensor modalities with the rest sensor modalities. The free-field microphones ($Mic_2$) were not included in this evaluation, as they had performed too poorly overall. 

Apart from comparing experimental results in terms of mean, standard deviations, and 95\% CI, we conducted statistical analyses. The performances of individual sensors-and sensor fusion-based models were first evaluated using the one-way analysis of variance (ANOVA) to test the hypothesis that there are statistically significant differences between multiple groups. Afterward, the best-performing model was compared to other models using a pair-wise t-test to verify if the best-performing model was significantly better than the counterpart. Similarly, a pair-wise t-test was further used to compare each transfer learning-based model with its corresponding individual sensor-based model and the performances of individual sensors-based models with respect to BMD. P values <.05 are considered to be statistically significant. 

\section{Results}

Experiment I showed that detection using contact microphones ($Mic_1$) showed the highest mean breach recall (85.8\%) compared to other sensors (Table \ref{Table:Fusion results}). In each table, the highest-performing sensor is indicated in bold. It was followed by another contact sensor ($PCB_1$) connected to the vertebra (81\% average breach recall, 77.9-84.1\% 95\% CI). In contrast, a free-field microphone performed the worst with 63.8\% average breach recall (59.1-68.5\% 95\% CI). Of the tri-axial accelerometer on the drill ($PCB_3$), the direction along the drilling axis (Z-axis) showed the highest performance (77.2\% average breach recall, 71.0-83.3\% 95\% CI).

The inference of each model was tested with respect to BMD. According to Table \ref{table:bmd test}, the results show significant differences in the models trained with $PCB_1$, $PCB_2$, and ${PCB}_{3x}$. $PCB_1$ and  ${PCB}_{3x}$ based models performed significantly better on the normal test set, whereas $PCB_2$ performed better on abnormal data (92.8\% average breach recall, 87.0-98.6\% 95\% CI).
\begin{table*}[ht]
\centering
\caption{Performances of the individual sensors- and sensors fusion-based models. P values are from the pair-wise t-test results with respect to $Mic_1$ \& $PCB_1$. }
\begin{tabular}{lccc}
\hline
Sensor & Breach Re. (95\% CI) [\%] & Non-breach Re.[\%] & P Value \\
 \hline
$Mic_1$	    & \textbf{85.8±3.19 (83.0-88.6)} & 91.8±2.60 & <.001\\ 
$Mic_2$       & 63.8±5.36 (59.1-68.5)	& 91.0±3.39 & <.001\\ 
$PCB_1$         & 81.0±3.54 (77.9-84.1)	& 87.6±7.02 & <.001\\ 
$PCB_2$	    & 79.8±8.65 (72.2-87.4)	& 92.0±0.71 & .002\\ 
${PCB}_{3x}$	        & 75.2±6.18 (69.8-80.6)	& 91.0±1.22 & <.001\\ 
${PCB}_{3y}$	        & 73.4±7.96 (66.4-80.4)	& 87.6±2.30 & <.001\\ 
${PCB}_{3z}$	        & 77.2±7.05 (71.0-83.3)	& 93.0±1.58 & <.001\\ 
\hline

Sensor Combination \\
\hline
 $Mic_1$ \& $PCB_1$  & \textbf{98.0±2.74 (95.6-100)} 	& 96.2±1.30& \\
$Mic_1$ \& $PCB_1$ \&  ${PCB}_{3z}$	     & 97.0±4.47 (93.1-100) 	    & 96.0±0.79 & .681\\ 
$Mic_1$ \& $PCB_1$ \& $PCB_2$    & 94.0±4.18 (90.3-97.7)	& 96.4±0.55 & .111\\ 
${PCB}_{3x}$ \& ${PCB}_{3y}$ \& ${PCB}_{3z}$       & 84.8±3.83 (81.4 -88.2) & 84.6±0.55 & <.001\\ 
$Mic_1$ \& $PCB_1$ \& $PCB_2$ \& ${PCB}_{3x}$	& 95.2±5.72 (90.2-100)	& 97.8±0.84 &.352\\ 
$Mic_1$ \& $PCB_1$ \& $PCB_2$ \&  ${PCB}_{3y}$    & 96.4±0.55 (97.0-100)	& 96.4±0.55& .545\\ 
$Mic_1$ \& $PCB_1$ \& $PCB_2$ \&  ${PCB}_{3z}$  & 96.0±2.24 (94.0-98.0)	& 96.9±0.89 &.242\\ 
$Mic_1$ \& $PCB_1$ \& $PCB_2$ \& ${PCB}_{3x}$\& ${PCB}_{3y}$ \& ${PCB}_{3z}$
 & 92.0±2.74 (89.6 - 94.4)	& 97.4±0.55 &.009\\ 

 \hline
 \end{tabular}
\label{Table:Fusion results}
\end{table*}

\begin{table*}[]\centering
\caption{Performances of individual sensors-based models with respect to the bone density. P values are from the pair-wise t-test between the results obtained from testing on normal and abnormal sets.}
\begin{tabular}{lccc}
 \hline
Sensor  & Normal - Breach Re. (95\% CI) [\%] &Abnormal - Breach Re. (95\% CI) [\%] & P value \\
 \hline
$Mic_1$  & 86.2±6.22 (80.7-91.6)               & 85.2±10.52 (76.0-94.4)  &  .859 \\
$Mic_2$ & 58.6±8.8 (50.9-66.3)                & 72.6±18.5 (56.4-88.8)  &   .165   \\
$PCB_1$   & 84.8±5.31 (80.1-89.5)             & 75±0.00 (75.0-75.0) &     .003             \\
$PCB_2$  & 72.6±11.7 (62.4-82.8)               & \textbf{92.8±6.57 (87.0-98.6)} &  .009             \\
${PCB}_{3x}$	     & \textbf{87.8±3.83 (84.4-91.2)}              & 55.4±11.24 (45.5-65.3)    & <.001            \\
${PCB}_{3y}$	   & 73.8±11.5 (63.7-83.9)               & 72.8±10.4 (63.7-81.9)    & .889             \\
${PCB}_{3z}$     & 77.0±8.0 (70.0-84.0)                & 77.8±16.1 (63.6-92.0)        & .923          \\

 \hline
\end{tabular}
\label{table:bmd test}
\end{table*}

The follow-up experiment II results,  investigating the power of transfer learning in the acoustic spectrogram domain, are presented in Table \ref{Table:transfer learning}. $PCB_1$  showed significantly improved performance with 93.0\% mean breach recall compared to the models trained from scratch. In contrast, using pre-trained weights on ${PCB}_{3y}$ deteriorated its average breach recall from 73.4\% to 62.8\%. Meanwhile, there were no significant differences in the case of other sensors, e.g., contact and free-field microphones. 
\begin{table*}
\centering
\caption{Performances of transfer learning-based models. P values are from the pair-wise t-test with respect to its corresponding model results trained from scratch.}
\begin{tabular}{lccc}
\hline
Sensor& Breach Re. (95\% CI) [\%] & Non-breach Re.[\%] & P value\\
 \hline

$Mic_1$	    &  85.6±4.51 (81.7-89.6)  & 89.6±1.66 & .937\\
$Mic_2$       & 60.0±5.70 (55.0-65.0)	& 88.2±1.08 & .309\\
$PCB_1$         & \textbf{93.0±2.74 (90.6-95.4)}	& 87.6±1.32 & <.001 \\
$PCB_2$	    & 79.0±2.74 (76.6-81.4)	& 92.6 ± 0.8 & .849\\
${PCB}_{3x}$		        & 74.0±2.74 (71.6-76.4)	& 91.2 ± 0.75 & .702 \\
${PCB}_{3y}$		        & 62.8±5.07 (58.4-67.2)	& 87.0±1.05 & .036\\
${PCB}_{3z}$	        & 83.8±2.33 (81.6-86.4)	& 94.3±1.47  & .079\\
 \hline
 
 \end{tabular}
 \label{Table:transfer learning}
\end{table*}

The main experimental results on sensor fusion were integrated into Table \ref{Table:Fusion results}. Eight different combinations were tested in total. We first combined the best-performing sensors ($Mic_1$ and $PCB_1$). We achieved the highest performance, particularly 98.0\% and 95.6-100\% 95\% CI. Then, we added consecutive sensor types in the next experiments to test the combination of three sensor modalities. As a result, adding more sensor data (e.g., $PCB_2$) deteriorated the sensitivity  (94.0\% average breach recall, 90.3-97.7\% 95\% CI). We also combined X-, Y-, and Z-directional data from $PCB_3$ and increased average breach recall up to 84.8\%. Concatenating 4 sensor data improved the breach recall by 2\% in the case of $Mic_1$ \& $PCB_1$ \& $PCB_2$ \&  ${PCB}_{3z}$ compared to $Mic_1$ \& $PCB_1$ \& $PCB_2$. However, combining all sensor data dropped the average breach recall by 4\% compared to $Mic_1$ \& $PCB_1$ \& $PCB_2$ \&  ${PCB}_{3z}$. Nevertheless, compared to individual sensor-based models, the performance of data fusion is higher: e.g., $Mic_1$, $PCB_1$, and $PCB_2$ show 85.8\%, 81.0\%, and 79.8\% average breach recall, while their fusion resulted in 94.0\%. Furthermore, we conducted a one-way ANOVA on all individual sensors and their combinations, which showed significant differences between the groups (Table \ref{Table:Fusion results}). Afterward, pairwise t-tests between the best combination ($Mic_1$ and $PCB_1$) with all individual sensors and their combinations were performed. As a result, the fusion of $Mic_1$ and $PCB_1$ was significantly better than all individual sensors-based models, and the combination of ${PCB}_{3x}$ \& ${PCB}_{3y}$ \& ${PCB}_{3z}$ and the combination of all sensors according to Table \ref{Table:Fusion results}.

\section{Discussion} 
This study investigated the potential of vibroacoustic sensors in detecting pedicle breaches. The contact between the drill and bone produces vibrations that vibroacoustic sensors can capture, and it has been proved that they can discern between cortical and cancellous layers of the bone~\cite{zakeri2017classifying,Guan2018,Sun2014,Torun2018,torun2020parametric}. In our work, we physically generated breaches controlled and accurately by integrating a surgical navigation system to translate the preoperative planning to the anatomy. Subsequently, we developed an automatic data labeling pipeline based on the tracking data obtained from the navigation setup. We explored the prospect of a contact microphone, a free-field microphone, a uni-axial accelerometer, and a tri-axial accelerometer enabled by the squeeze-and-excitation network (SE-ResNet-18) in detecting breaches. \par

Among all vibroacoustic sensing methods, contact microphones showed the highest sensitivity of 85.8±3.19\%, which was followed by the uni-axial accelerometer connected to the bone (81.0±3.54\%) (Table \ref{Table:Fusion results}). This tendency can be understood from Fig.~\ref{Fig: spectograms}. One can notice that spectrograms from contact microphones and uni-axial accelerometers attached to the bone and surgical pin provide more distinctive features for significant events. These sensors are the closest to the actual location of possible breach events. They are less susceptible to environmental noise than the free-field microphone, which performed poorly (63.8\% mean breach recall). The sound waves generated in the bone also had to travel ten to twenty centimeters through the air before reaching the free-field microphone. Regarding the axes of the tri-axial accelerometer, the signal from the Z-direction (along the direction of drilling) is the one performing strongest. This fact can be justified by its being a direction of signal propagation, thus providing stronger signals.

The evaluation revealed the differences in bone quality. Determining bone quality based on vibroacoustic signals is a promising topic for future research. In our opinion, state-of-the-art lacks methods estimating bone density. Therefore, we tested the networks with respect to BMD, characterized as normal and abnormal. Overall, no statistical significance could be shown except for uni-axial accelerometers and the tri-axial accelerometer in the X-direction. Regarding contact microphones, the average sensitivity was similar for both types, around 85\%. It shows that contact microphones are indifferent to bone quality for breach detection. Sensors showing significant differences between breach recall for normal and abnormal bone, such as the Z-axis of the tri-axial accelerometer (${PCB}_{3x}$), could be due to the prevalence of a lower signal-to-noise ratio in the abnormal test set. However, further investigations are necessary to define the cause and why some sensor configurations are more susceptible to the BMD.

Furthermore, we implemented the concept of transfer learning by using the weights trained on the breakthrough detection data in the same mel-spectrogram by \cite{seibold2021real}. It was used as an initialization for our models. We expected to see an added performance on the contact microphones since pre-trained weights are also trained on the data from contact microphones. However, statistical testing with respective models trained from scratch (Table \ref{Table:transfer learning}) did not reveal significant general differences except for the uni-axial accelerometer on the bone and the Y-axis of the tri-axial accelerometer. However, the trend for these two sensors is controversial.  \par
Apart from inputting individual sensor data to the network, their different combinations were examined to prove the hypothesis that sensor fusion boosts performance and robustness. Table~\ref{Table:Fusion results} shows how much the fusion of several sensor data improves the accuracy of the network. For most sensor combinations, the accuracy is pushed well above 90\% for both breach and non-breach classification. The high variability for some sensors in Table \ref{Table:Fusion results} is likely because the signal of breach events largely varies depending on the breach location. Due to differences in geometry, bone thickness, drill angle, and other parameters, a medial breach can cause a different response compared to a lateral breach. Thus, these variations perplex the network to generalize from the limited data. Owing to the direct contact of sensors with the tissue, the best-performing combination is contact microphones with the uni-axial accelerometer on the bone (98\% breach recall), which is the same as the performance of the PediGuard probe with fluoroscopic guidance  \cite{chaput2012reduction}. It shows that using vibroacoustic sensors with a deep learning algorithm is a reliable alternative to fluoroscopy-supported interventions to minimize radiation exposure. \par 

We believe the proposed method is a promising approach to navigate either hand-held or robotic surgical drilling to facilitate safe pedicle screw placement. However, this study still has several limitations before translation to a real clinical scenario. First of all, in this study it is the detection of breach events that is analyzed and what the network is trained for. In a clinical setting it is, however, most useful to predict the breach event to stop the drilling action before the drill breaches the pedicle wall. A next step in this research is to train a model to predict an imminent breach. Further, the results of this study are based on ex-vivo experiments. Before discerning the clinical feasibility, the developed technology should be tested on in-vivo animal experiments for a close-to-real setup. The number of samples was limited due to the time frame, efforts, and costs of performing cadaver experiments. The performance could be further improved by having more samples and investigating synthetic data generation techniques for CNN training. In terms of the direct translation of the sensor technology into the operating room, one of the current limitations is the non-sterilizable nature of the contact microphones and uni-axial accelerometers. Thus, sensors must be further developed to be sterilizable. Finally, we only analyzed the spectrograms over a frequency range from 0 to 2kHz. However, some investigated sensors might give better results in higher frequency ranges. According to ~\cite{Dai2020}, high-frequency components in tactile vibration feedback contain the most critical information about transient contact events. This study represents the first effort to translate non-visual sensing technology to the operating field as a local sensing method for hand-held and robotic orthopedic surgery applications.

\section{Conclusion} 

In this paper, we conducted the first ex-vivo study on a deep learning-driven non-vibroacoustic sensing approach for breach detection during pedicle screw placement in spine surgery. The non-visual nature of vibroacoustic sensing eliminates errors caused by non-negligible patient movements during operation and radiation exposure to the patient and medical staff. The results are similar to those of the PediGuard probe, the only non-visual sensing device on the market. The proposed method has great potential to increase the autonomy of robotic surgery by introducing non-visual physical intelligence. We also proposed a new method of automatic data labeling based on surgical navigation, which can be beneficial in labeling a large amount of data in the future.

\bibliographystyle{IEEEtran}
\bibliography{main.bib}
\end{document}